\documentclass[acmsmall,nonacm]{acmart}

\usepackage[T1]{fontenc} 
\usepackage[utf8]{inputenc}

\usepackage{graphicx}

\usepackage[]{hyperref}
\hypersetup{
    colorlinks=false,
}

\usepackage{subfigure}
\usepackage[squaren]{SIunits}
\usepackage{booktabs}

\setcitestyle{unsrt}

\newcommand{\ra}[1]{\renewcommand{\arraystretch}{#1}}

\newcommand{\rsq}{$\text{R}^2$}
\newcommand{\model}{MapLUR}

\newcount\ppnum
\newcommand\ppnumber[1]{%
        \ppnum=#1\relax
        \ifnum\ppnum<0
                $-$%
                \ppnum=-\ppnum
        \fi
        \let\pptemp\empty
        \loop\ifnum\ppnum>999
                \count255=\ppnum
                \divide\ppnum by1000
                \count255=\numexpr \count255 - 1000*\ppnum \relax
                \edef\pptemp{,\ifnum\count255<100 0\ifnum\count255<10 0\fi\fi
                                \the\count255 \pptemp}%
        \repeat
        \the\ppnum
        \pptemp
}

\newcommand{\num}[1]{\ppnumber{#1}}
\newcommand{\SI}[2]{#1 #2}
\newcommand{\si}[1]{#1}

\newcommand{\NOtwo}{$\text{NO}_2$}

\newcommand{\PMtwofive}{$\text{PM}_{2.5}$}
\newcommand{\PMten}{$\text{PM}_{10}$}

\newcommand{\baselines}{
    Mean baseline & 0.000 & 13.971 \\
    Linear regression & 0.487 & 10.004 \\
    Multi-layer Perceptron & 0.499 & 9.887 \\
    Random Forest   & 0.662  & 8.119  \\} 

\acmJournal{TSAS}

\begin{document}

\SIunits[]

\title{\model{}: Exploring a new Paradigm for Estimating Air Pollution using Deep Learning on Map Images}

\author{Michael Steininger}
\affiliation{%
    \institution{University of W{\"u}rzburg}
    \department{Institute of Computer Science, Chair of Computer Science X}
    \city{W{\"u}rzburg}
    \country{Germany}
}
\email{steininger@informatik.uni-wuerzburg.de}
\author{Konstantin Kobs}
\affiliation{%
    \institution{University of W{\"u}rzburg}
    \department{Institute of Computer Science, Chair of Computer Science X}
    \city{W{\"u}rzburg}
    \country{Germany}
}
\email{kobs@informatik.uni-wuerzburg.de}
\author{Albin Zehe}
\affiliation{%
    \institution{University of W{\"u}rzburg}
    \department{Institute of Computer Science, Chair of Computer Science X}
    \city{W{\"u}rzburg}
    \country{Germany}
}
\email{zehe@informatik.uni-wuerzburg.de}
\author{Florian Lautenschlager}
\affiliation{%
    \institution{University of W{\"u}rzburg}
    \department{Institute of Computer Science, Chair of Computer Science X}
    \city{W{\"u}rzburg}
    \country{Germany}
}
\email{lautenschlager@informatik.uni-wuerzburg.de}
\author{Martin Becker}
\affiliation{%
    \institution{Stanford University}
    \department{Stanford Medicine, Nima Aghaeepour Lab}
    \city{Stanford}
    \country{USA}
}
\email{mgbckr@stanford.edu}
\author{Andreas Hotho}
\affiliation{%
    \institution{University of W{\"u}rzburg}
    \department{Institute of Computer Science, Chair of Computer Science X}
    \city{W{\"u}rzburg}
    \country{Germany}
}
\email{hotho@informatik.uni-wuerzburg.de}

\renewcommand{\shortauthors}{Steininger et al.} 

\begin{abstract}
Land-use regression (LUR) models are important for the assessment of air pollution concentrations in areas without measurement stations. While many such models exist, they often use manually constructed features based on restricted, locally available data. Thus, they are typically hard to reproduce and challenging to adapt to areas beyond those they have been developed for.

In this paper, we advocate a paradigm shift for LUR models:
We propose the \textbf{D}ata-driven, \textbf{O}pen, \textbf{G}lobal (DOG) paradigm that entails models based on purely data-driven approaches using only openly and globally available data.
Progress within this paradigm will alleviate the need for experts to adapt models to the local characteristics of the available data sources and thus facilitate the generalizability of air pollution models to new areas on a global scale.

In order to illustrate the feasibility of the DOG paradigm for LUR, we introduce a deep learning model called \model{}.
It is based on a convolutional neural network architecture and is trained exclusively on globally and openly available map data without requiring manual feature engineering.
We compare our model to state-of-the-art baselines like linear regression, random forests and multi-layer perceptrons using a large data set of modeled \NOtwo{} concentrations in Central London.
Our results show that \model{} significantly outperforms these approaches even though they are provided with manually tailored features.

Furthermore, we illustrate that the automatic feature extraction inherent to models based on the DOG paradigm can learn features that are readily interpretable and closely resemble those commonly used in traditional LUR approaches.
\end{abstract}



\begin{teaserfigure}
   \includegraphics[width=\textwidth]{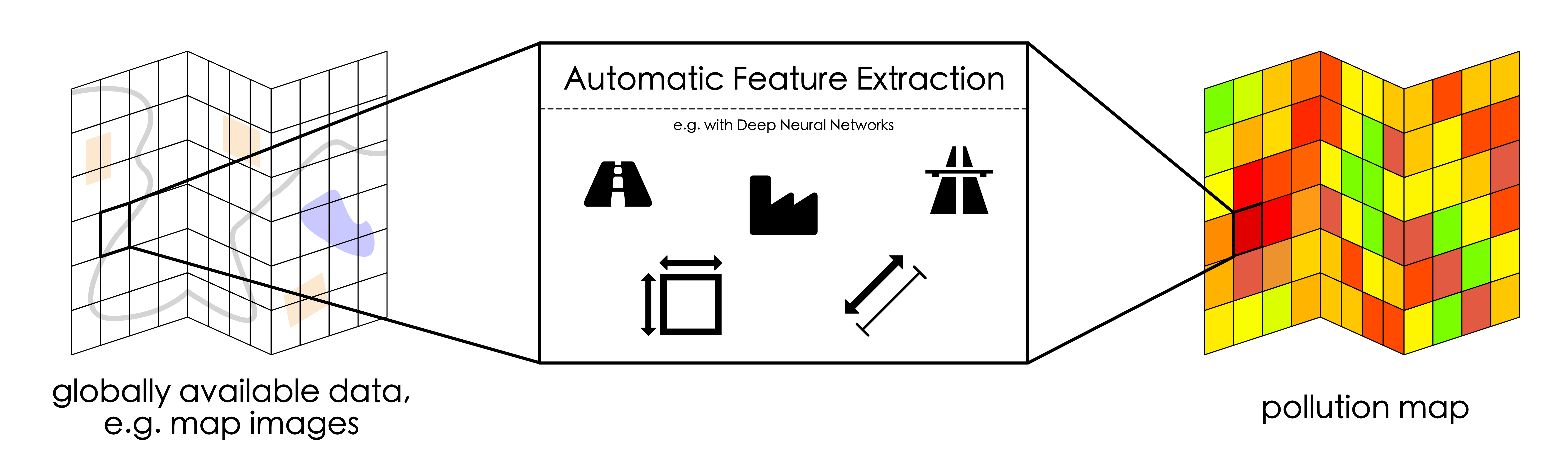} 
   \Description{The deep neural network automatically extracts air pollution features like roads or industry from map images. The model can be used to build pollution maps by providing it with map images. Based on these images, it estimates the air pollution concentration for the depicted locations using the learned features.}
   \caption{\textbf{\model{}: Automatic feature extraction and globally available data for air pollution modeling.} 
   We propose the DOG paradigm (Data-driven, Open, Global) for land-use regression which advocates openly and globally available data, and automatically extracting features in order to estimate pollution.
   Following this newly introduced paradigm, we propose the \model{} model. It consists of a deep neural network architecture that estimates pollution concentrations for specific locations directly from globally available map images (rendered maps or satellite images) resulting in area spanning pollution maps. For this, \model{} automatically learns to extract features from the given map images. The extracted features closely resemble manually engineered features for land-use regression models.
   }
   \label{fig:teaser}
\end{teaserfigure}

\maketitle

\section{Introduction}

Air pollution is known to have adverse effects on human health and
the environment~\cite{brunekreef2002air,elsom1992atmospheric}.
Thus, especially in areas with high population counts, it is important to control local pollution concentrations.
For this reason, monitoring stations are deployed in many cities, which measure pollution
continuously in order to assess whether the pollution levels are still within acceptable/legal limits.
However, since the number of stations in a city is usually very limited, there are many areas where no air quality data is available.
To fill these gaps, land-use regression (LUR) models are often used to
estimate pollution concentrations in areas without monitoring
stations~{\cite{hoek2008review,beelen2013development,wu2015applying}.

\paragraph{Problem Setting}
In recent years, a wealth of different land-use regression models have been developed
that have shown to provide promising pollution estimates.
However these models
    i) are at least partially based on neither globally nor openly available data~\cite{beelen2013development,hoek2015land,alam2015exploring} and
    ii) often rely on hand-crafted features.

Thus, due to the local nature of the features, 
    i) these models usually do not generalize easily to locations other than the one they were developed for.
    Additionally, due to the involved hand-crafting process, ii) optimizing the features for new models in specific study areas is a cumbersome process.

\paragraph{Approach}
To address the challenges inherent to inaccessible data and manual feature engineering in land-use regression models, in this work, we advocate a paradigm shift towards purely data-driven land-use regression models based on open and globally available data.
We call the corresponding paradigm \textbf{DOG} (\textbf{D}ata-driven, \textbf{O}pen, \textbf{G}lobal).
More specifically, models adhering to DOG work directly on raw data, automatically extracting their features from the input.
While such data-driven methods have proven successful in multiple application domains~\cite{lecun2015deep}, they have so far not been introduced to land-use regression.
Land-use regression models following the DOG paradigm have multiple advantages: 
i) they can be fit more easily to different study areas than other, more specialized land-use regression approaches,
ii) they do not require manual feature engineering, and 
iii) they can be reproduced by other researchers without requiring access to data sources that are not easily available.
In order to demonstrate the feasibility of this paradigm, we introduce the \model{} model.
\model{} implements DOG by using deep learning, specifically a convolutional neural network architecture. It automatically extracts features from map images, which are openly available almost anywhere on earth, and estimates air pollution based on these features.

\paragraph{Experimental Evaluation}
We assess the performance of \model{} by comparing it against state-of-the-art land-use regression models like linear regression, Random Forests (RF), and Multi-layer Perceptrons (MLP) on modeled \NOtwo{} concentration data
from the London Atmospheric Emissions Inventory (LAEI)~\cite{laei}.
In the process, we employ different types of images
including map images from OpenStreetMap and Google Maps~\cite{google2018maps} as
well as satellite imagery from Google Maps.
We find that our model works best using map images from OpenStreetMap and that it outperforms all baselines significantly.

Furthermore, we analyze the data requirements of \model{} and the baselines. We find that common for deep learning models, \model{} requires more training data than models that rely on hand-crafted features. In this context, we evaluate how far the training set can be reduced and discuss possible approaches to further address this challenge.

Finally, we analyze the automatically extracted features by observing which parts of the map images were particularly important for our model using guided backpropagation~\cite{springenberg2014striving} and artificial map images.
The analysis shows that the learned \model{} features strongly relate to hand-crafted features as commonly used in land-use regression models.

\paragraph{Contribution}

Our core contributions in this work are threefold:

\begin{enumerate}
    \item We propose DOG, a new, data-driven paradigm to land-use regression. Models following this paradigm should not require manual feature engineering and only rely on openly and globally available data sources.
    \item We introduce \model{}, a land-use-regression model based on DOG. \model{} employs a deep learning approach to automatically extract features from map images. We show that this model is able to outperform traditional land-use regression models when trained on a sufficiently large data set.
    \item We demonstrate that, contrary to popular believe, models based on the data-driven paradigm are not necessarily black-boxes by inspecting the features \model{} extracts, finding that the automatically extracted features strongly relate to typical manually engineered features for land-use regression models.
\end{enumerate}

\paragraph{Structure}
This work is organized as follows.
Related work is summarized in Section~\ref{sec:related_work}.
Section~\ref{sec:materials} describes the air pollution data and image data used in this work.
DOG and the \model{} model are introduced in Section~\ref{sec:methods}.
The experiments and the baseline models are described in Section~\ref{sec:evaluation}.
Section~\ref{sec:results} presents our results and analyzes our model.
We discuss advantages and limitations of DOG and \model{} in Section~\ref{sec:discussion}.
Finally, Section~\ref{sec:conclusion} concludes this work.

\section{Related Work}
\label{sec:related_work}

Land-use regression has been an active field of research for many years now.
Work done in the previous decade has laid important foundations for current land-use
regression models and established linear regression techniques as the de facto standard
model~\cite{beelen2013development,wu2015applying,muttoo2018land}.
Especially noteworthy is the Escape project~\cite{eeftens2012development,beelen2013development} which built models for \num{36} European areas.
The model building procedure of this project has become a standard approach~\cite{muttoo2018land,wolf2017land,meng2015land,montagne2015land,wang2014performance}.
In order to make the application of land-use regression models easier, there is a tool available which automizes the process of variable generation, modeling and prediction with a model based on linear regression~\cite{morley2018land}.

However, more advanced machine learning methods are starting to become more common.
One example for these approaches are Random Forests~\cite{breiman2001random}.
They have been used successfully to estimate elemental components of particulate matter in Cincinnati, Ohio~\cite{brokamp2017exposure} and \NOtwo{} pollution in Geneva~\cite{champendal2014air}.

Another example for a more advanced method are neural networks.
These models have been used to estimate a range of pollutants successfully, as shown in various publications.
For example, they have been applied to \NOtwo{}~\cite{adams2015advancing,liu2015land,champendal2014air}, \PMtwofive{}~\cite{adams2015advancing,fan2017spatiotemporal,xie2018autoencoder,bai2019ensemble,bai2019hourly}, \PMten{}~\cite{alam2015exploring,liu2015land,xie2018autoencoder}, and surface dust~\cite{buevich2016modeling} concentrations.
The neural-network-based models are typically simple multi-layer Perceptrons.
However, there are deep learning models which use recurrent neural networks or deep belief regression networks.
These models differ from this work in that they are used to forecast pollution concentrations from earlier measurements or fill missing values for locations where measurements already exist~\cite{fan2017spatiotemporal,xie2018autoencoder,bai2019ensemble,bai2019hourly}, while we estimate pollution for locations without measurements.
To the best of our knowledge, there are no deep learning models for our setting.
Both Random Forests and neural networks have been shown to outperform linear regression in land-use regression~\cite{champendal2014air,brokamp2017exposure}.

Support vector regression models~\cite{drucker1997support} are another possible approach.
There are models which can forecast pollution concentrations using this technique~\cite{ortiz2010prediction,li2019air}, but there do not seem to be any land-use regression models with this type of model.

All aforementioned land-use regression models rely on manually
engineered features, which are typically gathered from various locally
available data sources that might not be available elsewhere.
In contrast to all methods above, we propose a deep learning model based on
convolutional neural networks (CNNs), which is able to automatically learn relevant features from openly available maps.

Such image-based approaches have been used before in the context of air quality estimation and pollution detection.    
\citet{singh2016higher} interpreted modeled air pollution data as images and used non-machine-learning image classification techniques in order to detect higher pollution episodes.
Furthermore, CNNs have been used before in the context of air quality estimation by \citet{zhang2018end} and \citet{li2015using}, who proposed models to estimate air haze level using photos from, for example, mobile phones or webcams.
In contrast, our work uses map and satellite imagery depicting land-use as model input, making our model more closely related to land-use regression models. Additionally, our model estimates pollution concentrations instead of haze levels.

\section{Materials}
\label{sec:materials}

In this section, we introduce the air pollution data set we use to train and evaluate
our method as well as the data sources from which we extract map and satellite
images.

\subsection{Air Pollution Data}
\label{sec:air_pollution_data}

We train and test our model using pollutant concentrations from the London Atmospheric Emissions Inventory (LAEI)~\cite{laei}.
It contains modeled annual mean concentrations of \NOtwo{} and \PMten{}, among other pollutants, at a \SI{20}{\meter} grid level for the complete Greater London area in 2013.
For our main model development and evaluation we use the \NOtwo{} concentrations of
the data set since it is a very frequently used pollutant for land-use regression models.
The data is the result of a dispersion model which incorporates a vast number of input factors like for example road and rail networks, traffic data, aviation, pollution from individual industrial premises, domestic and commercial fuel consumption, as well as fires.
Through this approach, \num{5856428} data points were generated where each data point represents a \SI{20}{\meter} by \SI{20}{\meter} cell~\cite{laei}.

We sample a training set consisting of \num{3000} data points and a test set consisting of \num{1500} data points from the Central London part of the data set in order to have a reasonable number of urban data points for our experiments.
We choose data points from Central London because we believe that it is more important in practice to reliably estimate pollutant concentrations in highly polluted areas with a large population than in more rural areas.
For this, we define a geographical rectangle that roughly describes Central London and only use cells within.
The box's north western corner is at \num{526660} easting and \num{183220} northing while the south eastern corner is at \num{534760} easting and \num{177640} northing specified in British National Grid coordinates.
The sampled data points are depicted in Figure~\ref{fig:data_split} in Appendix~\ref{sec:appendix_laei_cells}.
Descriptive statistics for the cells in Central London and the sampled subset used for training and testing can be found in Table~\ref{tab:laei_no2}.

\begin{table}
    \centering
    \ra{1.3}
    \caption{\textbf{Statistics of London's \NOtwo{} concentrations.} The statistics for Central London include all LAEI cells in the bounding box. The sampled data set contains randomly sampled cells from the Central London data set, which are used for training and testing models in our experiments. Mean, standard deviation (SD), minimum values (Min), and maximum values (Max) are in \si{\micro\gram\per\cubic\meter}.}
    \begin{tabular}{@{}lccccc@{}}
    \toprule
    Data set & Count & Mean & SD & Min & Max \\
    \midrule
    Central London & \num{113680} & 50.90 & 15.02 & 37.12 & 253.89 \\
    Sampled subset & \hspace{3.5mm}\num{4500} & 50.85 & 15.02 & 37.17 & 171.06 \\
    \bottomrule
    \end{tabular}
    \label{tab:laei_no2}
\end{table}

The map images, which we use to depict the areas of the data points, show \SI{80}{\meter} by \SI{80}{\meter} areas even though the air pollution data is available at a \SI{20}{\meter} grid level.
The \SI{20}{\meter} by \SI{20}{\meter} cells are in the center of these images.
This allows \model{} to see more of the surroundings and incorporate information about distant emission sources.
In order to avoid a potential evaluation issue, we sample data points in such a way that no images can overlap.
Any overlap could lead to a situation where the model already roughly knows the pollution concentration for a test data point since it might recognize the test data point's area from the image of a nearby training data point.
Such implicitly learned proximity of data points could give our model an unfair advantage, which we avoid with our procedure.

\subsection{Image Data}
This section describes the sources of map and satellite image data that we use in this paper as well as the preprocessing applied to the images in order to generate training samples.

\subsubsection{Image Sources}
\label{sec:image_sources}
There is a variety of globally available sources for map images, two
popular services being Google Maps~\cite{google2018maps} and
OpenStreetMap~\cite{OpenStreetMap}.
While Google Maps is a commercial and proprietary service, OpenStreetMap is an
open database for map data that is built and maintained by volunteers.
Data from OpenStreetMap can be used to render maps in various ways through
different stylesheets.
In this work, we render map images based on OpenStreetMap data using a slightly
modified version of the default stylesheets used on the official OpenStreetMap
website.
It differs from the default in that we do not render text like street or station
names, since labels obstruct map features making them harder to recognize and often only carry very localized information, thus, possibly reducing generalizability.
For tile rendering we use \emph{mod\_tile} which is a module for the Apache web
server with the rendering back-end \emph{renderd}~\cite{modtile}.
In addition to OpenStreetMap images, we use map and satellite images
from Google Maps, in order to compare the effectiveness of each visualization for
this task.
Since Google Maps is proprietary, the images cannot be easily customized to the same extent as OpenStreetMap images.
We therefore use them without modification.

\subsubsection{Image Preparation}
\label{sec:image_preparation}
Before using \model{} it is necessary to prepare map or satellite images.
We found through preliminary experiments that images depicting \SI{80}{\meter} by \SI{80}{\meter} provide the best performance in this setting, as can be seen in Appendix~\ref{sec:appendix_area_size}.
To depict the correct area in an image we approximate the \SI{80}{\meter}
distances using a meter per pixel value that depends on
the zoom level of the image and the latitude of the location due to the
Mercator projection, which both OpenStreetMap and Google Maps use.
We obtain the images at zoom level \num{17}, resulting in an pixel extent of approximately \SI{0.75}{\meter} by \SI{0.75}{\meter} at London's latitude.
Thus, the images have a resolution of 106 px by 106 px.
The rendered images are then scaled to a fixed resolution of 224 px by
224 px similar to what popular CNN architectures for the ImageNet
competition~\cite{imagenet_cvpr09} use.
This way we avoid having to change the model architecture when depicting
different sized areas in the images or when we fit the model to locations with a
different latitude which would also result in differently sized images.
Using a model input resolution that fits the images exactly could reduce the
model size and improve training and inference speed. Nonetheless, we found it unnecessary
considering the already acceptable speed and we favored the increased
flexibility.

\section{Methods}
\label{sec:methods}
In the following, we introduce our data-driven paradigm DOG to
land-use regression, which suggests that models should estimate air pollution by
automatically extracting relevant features from openly and globally available data.
We also present our model \emph{\model}, which follows this paradigm by using a convolutional
neural network as an automatic feature extractor, taking as input globally and openly available map images.

\subsection{The DOG Paradigm}

Our data-driven paradigm DOG (Data-driven, Open, Global) aims to alleviate the issues of manual feature engineering as well as only locally applicable models.
To this end, it requires models to fulfill the following criteria:

\begin{itemize}
    \item \textbf{Automatic extraction of relevant features}:
        Models should learn a function from the raw input to the desired output, which leads to automatic development of features and may even uncover relevant factors that are not yet known as having an influence on air pollution.
        This can for example be achieved by deep learning methods like the convolutional neural network introduced in the next section.
    \item \textbf{Usage of globally available data sources}:
        Models should rely exclusively on data sources that are available for (almost) all parts of the world.
        This allows the ubiquitous application of the model without collecting additional, only locally available data sources.
    \item \textbf{Usage of openly available data sources}:
        Models should rely exclusively on openly available data sources.
        This allows researchers to reproduce and improve the model's results without requiring access to paid or
        not publicly available resources.
\end{itemize}

\subsection{The \model{} Model}
\label{subsec:model}
In this section, we propose the specific model \model{} based on the DOG paradigm described above.
Our model applies the paradigm by using map images as a globally, openly available source of information and extracting features through deep learning, more specifically a convolutional neural network (CNN).
This type of network is a natural fit for our setting, since it is specifically designed to work with two-dimensional shapes like images.
CNNs utilize spatial locality within the images and reduce the number of learned parameters through weight sharing.
These concepts make it feasible to learn relevant features from raw images, in contrast to fully-connected networks, which would need an impractical amount of parameters~\cite{lecun1998gradient}.

The structure of \model{} is depicted in Figure~\ref{fig:model}.
It contains \num{15} convolutional layers with batch
normalization~\cite{ioffe2015batch} and rectified linear units
(ReLUs), which are pair-wise linear activations~\cite{nair2010rectified}.
The last convolutional layer is followed by a fully connected layer with
\num{128} neurons and ReLU activation (depicted as the third and second to last boxes in Figure~\ref{fig:model}).
These neurons are then connected to a single neuron with linear activation that
produces the estimated pollution concentration.
Each convolutional layer has \num{16} filters, a kernel size of \num{3}, a
padding of \num{1}, a dilation of \num{1}, and a stride of \num{1}.
The output size of these layers is the same as their input size.
Maximum pooling layers with a kernel size of \num{2} and stride of \num{2} are
applied after the ReLUs of the first, third, fifth, seventh, tenth, and
thirteenth convolutional layer in order to reduce the number of activations.
We found this architecture and the corresponding hyperparameters by evaluating different variations of the model using ten fold cross-validations on the training set.

We use ReLU activations since they have shown to work well for many different tasks and models, making them the most popular activation function for deep learning applications~\cite{lecun2015deep}.
While trying different architectures we have also experimented with SELU~\cite{klambauer2017self} and RReLU~\cite{xu2015empirical} activations but we found no improvements with these functions.
The linear activation in the final layer is common practice for regression tasks~\cite{lathuiliere2019comprehensive}.
It does not restrict the range of resulting values allowing the model to estimate any value.

\begin{figure}
    \centering
    \includegraphics[]{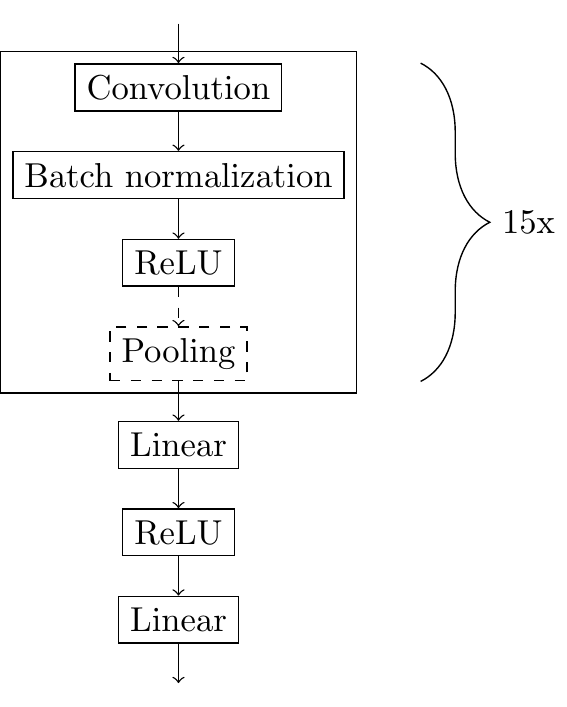}
    \Description{The structure of \model{}. It contains \num{15} convolutional layers with batch normalization and rectified linear units (ReLUs). The last convolutional layer is followed by a fully connected layer and ReLU activation. These neurons are then connected to a single neuron with linear activation that produces the \changed{estimated} pollution concentration.}
    \caption{\textbf{Structure of \model{}.} The model consists of \num{15} feature-learning building blocks which contain a convolution layer, batch normalization, rectified linear units (ReLUs), and sometimes a pooling layer. These building blocks are concatenated and only the first, third, fifth, seventh, tenth, and thirteenth block contain a pooling layer. These blocks are followed by a simple fully connected layer with ReLU activation and finally a single fully connected neuron with linear activation which returns the estimation of the pollution at the given location.}
    \label{fig:model}
\end{figure}

\section{Evaluation}
\label{sec:evaluation}
In order to evaluate \model{}, we conduct several experiments and compare our model to baseline models which are commonly used in land-use regression.
The experiments and the baseline models are described in the following.

\subsection{Experimental Setting}

We conduct four experiments using \model{}, varying the data available to the model.
For all experiments, \model{} is trained using the Adam optimizer~\cite{kingma2014adam} on batches of size \num{400} for at most \num{2000} epochs with a learning rate of 0.0001.
We augment the training data by flipping or transposing the images.
Additionally, we employ early stopping, interrupting training when the validation performance has not increased for \num{20} epochs in a row.

\paragraph{Experiment 1 --- OpenStreetMap.}
In the first experiment, only OpenStreetMap images are used as input to the CNN.
The input images have three channels (RGB), \SI{224}{px} by \SI{224}{px}, and depict an area of
\SI{80}{\meter} by \SI{80}{\meter}.
All labels were removed from the rendering process of the images, as described in Section~\ref{sec:image_sources}.

\paragraph{Experiment 2 --- Google Maps.}
Instead of OpenStreetMap images, Google Maps images were captured and fed into the model in this experiment.
The same size as in the previous experiment was used.
As Google Maps data is proprietary, modifications cannot be made as easily and to the same extent as with OpenStreetMap images.
Therefore, text labels are present in the imagery.

\paragraph{Experiment 3 --- Google Maps Satellite.}
For the third experiment, instead of stylized map images, we use satellite
images from Google Maps Satellite~\cite{google2018maps} which uses imagery
from both satellites and aerial surveys.
Training and test images from Google Maps Satellite have the same size and zoom levels as the OpenStreetMap and the Google Maps images.

\paragraph{Experiment 4 --- OpenStreetMap and Google Maps Satellite.}
Experiment \num{4} then combines OpenStreetMap images and satellite images by concatenating the two RGB images to one six-channel tensor.

\subsubsection{Evaluation Setup}
The models are evaluated using standard metrics for the evaluation of land-use regression models, namely \rsq{} and RMSE.
Both metrics are explained in Appendix~\ref{sec:appendix_evaluation_metrics}.
In all experiments, the model is initialized and trained \num{40} times on the
training set and evaluated on the test set, both of 
which are described in Section~\ref{sec:air_pollution_data}.
The average of the resulting evaluation metrics is then used as the final score
to counteract unfortunate initialization results.
Additionally, the sample of \num{40} evaluation runs can be used as the input to
statistical significance tests to formally confirm differences in evaluation results.

\subsection{Baselines}
In order to determine how well our model works, it is necessary to compare it to other methods.
Therefore, we first describe a set of features that is used by our baseline models.
Thereafter, four baseline models are introduced, namely a mean baseline, linear regression, Random Forest, and multi-layer Perceptron.
The last three of the aforementioned models are commonly used in land-use regression.
Random Forests and multi-layer Perceptrons tend to yield state-of-the-art results as described in Section~\ref{sec:related_work}.

\subsubsection{Features}
We use a set of standard land-use and road-related features for our baseline models.
These features have
shown to be important influencing factors for air pollution~\cite{eeftens2012development}.
All of these features can be calculated from OpenStreetMap data, since we want 
to provide similar information to all models for a fair comparison.
The features include the areas of commercial, industrial, and residential land-use, the lengths of big and local streets, and the distances to the next traffic signal, motorway, primary road, and industrial premise.
Big streets include streets that are classified as either motorway, trunk road, primary road or secondary road in OpenStreetMap while all other streets are local streets.
Most features are typically calculated for different \emph{buffers}, which are
areas with a specific radius around data points.
For example, the areas of different types of land-use and the lengths of streets
are calculated for \SI{50}{\meter} and \SI{100}{\meter} buffers in order to give
the baseline models similar sight into the surroundings as the \model{} model.
However, the features which calculate the distance from each
data point to specific locations like the closest traffic signals, roads or
industrial premises exist only once and are not calculated for different buffers.
Due to these features, the baseline models are given a slight advantage since they can get
information from entities which are further away than \model{} can see.

\subsubsection{Mean}
A simple baseline for a regression task is the mean baseline.
It disregards all features and estimates the mean value of all training data points for each test data point.
This baseline provides performance values that every other model should beat.

\subsubsection{Linear regression}
The most common approach to land-use regression is linear regression.
Therefore, it is useful to compare our novel model to this type of model.

We use the same supervised stepwise selection as \citet{eeftens2012development} for selecting the most relevant subset of features.
A description of this procedure can be found in Appendix~\ref{sec:appendix_model_building_linear_regression}.
After applying the stepwise selection on the development set the model is left
with the variables \emph{length of big streets} (\SI{50}{\meter} buffer),
\emph{distance to the next industrial premise}, and \emph{distance to the next
traffic signal}.

\subsubsection{Random Forest}
The Random Forest is a more powerful model that was shown to work well for land-use regression and can often provide better performance than typical linear regression approaches, as it can model non-linear correlations between features~\cite{brokamp2017exposure,champendal2014air}.
Therefore, we use it as another baseline in this work.

This model is built in a similar way to the procedure
in~\citet{brokamp2017exposure}.
Details are in Appendix~\ref{sec:appendix_model_building_random_forest}.
The final Random Forest model uses the variables \emph{distance to the next
industrial premise}, \emph{distance to the next primary road}, \emph{distance to
the next traffic signal}, \emph{distance to the next motorway}, \emph{length of
big streets} (\SI{50}{\meter} buffer), and \emph{area of residential land-use}
(\SI{100}{\meter} buffer).
It builds \num{394} trees using bootstrap samples, considers at most
42.79\,\% of the available features per split, needs at least three
samples to split a node, and needs at least three samples for a leaf node.

\subsubsection{Multi-layer Perceptron}
Neural networks, or more specifically multi-layer Perceptrons (MLPs), are models whose popularity for land-use regression tasks has grown in recent years and which often outperform other baselines~\cite{buevich2016modeling,adams2015advancing,alam2015exploring,liu2015land,champendal2014air}.
Additionally, evaluating multi-layer Perceptrons (even if not directly applied to image data) illustrates the performance of neural networks which, in contrast to the \model{} model, are not based on convolutions.

Again, we follow the model development procedure of previously published work.
In this case, we base our procedure on the one used by~\citet{alam2015exploring}.
Appendix~\ref{sec:appendix_model_building_mlp} contains a description of this procedure.

The MLPs use the variables \emph{distance to the next industrial premise}, \emph{distance to the next primary road}, \emph{distance to the next traffic signal}, \emph{distance to the next motorway}, \emph{length of big streets} (\SI{50}{\meter} buffer), \emph{length of local streets} (\SI{100}{\meter} buffer), \emph{area of industrial land-use} (\SI{100}{\meter} buffer), \emph{area of commercial land-use} (\SI{100}{\meter} buffer), and \emph{area of residential land-use} (\SI{100}{\meter} buffer).
The architecture search found that the best performing MLP model has a single hidden layer with \num{29} neurons.
This is similar to the MLPs used by previous publications~\cite{buevich2016modeling,adams2015advancing,alam2015exploring,liu2015land,champendal2014air}.

\section{Results and Analysis}
\label{sec:results}

Given the baseline methods and \model{}'s description, we now present the results for our experiments and analyze \model{} in terms of data requirements and features learned.

\subsection{Experiments}
Table~\ref{tab:results} shows the results of the baseline methods as well as \model{}'s results for our experiments.
All results in the Table are significantly different from each other.
To verify this, the metrics of each model are tested for normality using the test from D'Agostino and Pearson~\cite{d1971omnibus,d1973tests} with $p < 0.05$.
The statistical significance for models whose metrics are normally distributed are tested using a t-test, while the other models' metrics are tested with the Wilcoxon signed-rank test, both testing for $p < 0.05$.
Additionally, Bonferroni correction~\cite{bonferroni1936teoria} is applied, which further substantiates the statistical significance, since $p < \frac{0.05}{n}$ with $n = 7$ for each model pair.
$n = 7$ is chosen to account for the number of hypotheses that are tested on the same data (each model is tested against 7 other models). 

\paragraph{Baselines}
As described before, we use a simple mean baseline, a linear regression, an approach with Random Forests, and a multi-layer Perceptron with manually engineered features from OpenStreetMap data.
The results in Table~\ref{tab:results} show that the Random Forest is performing considerably and significantly better than both the linear regression and the MLP.

\paragraph{Experiment 1 --- OpenStreetMap}
Our model with OpenStreetMap images performs better than all baselines regardless of metric, which can be seen in Table~\ref{tab:results}.

\begin{figure}
    \centering
    \subfigure[LAEI data]{\includegraphics[width=0.4\textwidth]{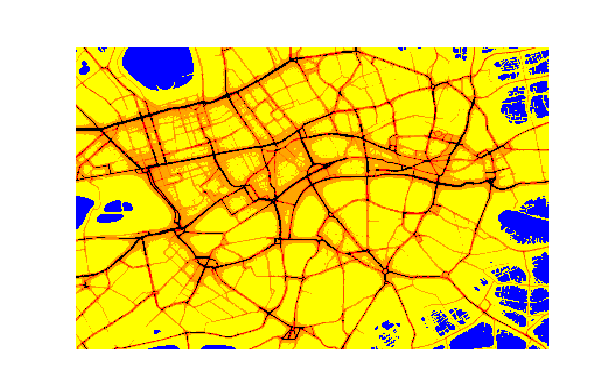}}
    \subfigure[\model{} estimates]{\includegraphics[width=0.4\textwidth]{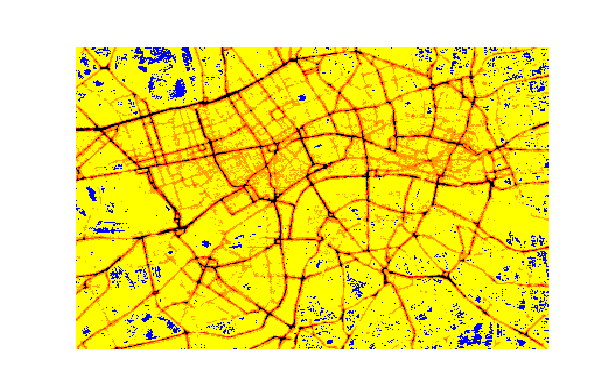}}
    \subfigure{\includegraphics[width=0.1\textwidth]{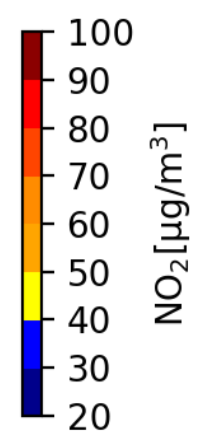}}
    \Description{The left image visualizes the pollution in Central London using the LAEI data while the right image shows the pollution in Central London based on \model{}'s estimates. The images look similar but \model{} tends to overestimate values of areas with low pollution concentrations and underestimate values of areas which are not in the vicinity of roads.}
    \caption{\textbf{Comparison of LAEI data and \model{} estimates.} The estimated map shares strong similarities with the original data. It can be seen that \model{} is able to recognize streets and accurately associate them with high pollution concentrations. However, our model tends to overestimate pollution in areas with very low pollution concentrations and underestimate pollution for areas with no road close by.}
    \label{fig:prediction_map}
\end{figure}

Figure~\ref{fig:prediction_map} shows the original \NOtwo{} concentrations of the LAEI data set in Central London and the estimates of \model{}.
It can be seen that our model is able to come rather close to the original data using only OpenStreetMap images, but tends to overestimate values of areas with low pollution concentrations and underestimate values of areas which are not in the vicinity of roads.

\paragraph{Experiment 2 --- Google Maps}
This experiment uses Google Maps imagery instead of OpenStreetMap images.
A drop in \rsq{} of more than \num{10} percentage points in comparison to the previous experiment and a higher RMSE value may be explained by the styling of Google Maps images.
Googles Maps contain fewer color-coded entities.
Especially streets, that are a common entity for land-use regression features, are not diversified as much as in OpenStreetMap images.
The differences can be seen in Appendix~\ref{sec:appendix_comparison_osm_gmaps}.

\paragraph{Experiment 3 --- Google Maps Satellite}
This experiment uses Google Maps Satellite imagery as input.
Table~\ref{tab:results} shows that using only Google Maps Satellite imagery leads to a considerable drop in performance, even worse than the linear regression baseline with an \rsq{} of 0.206 and an RMSE of 12.389.

These results are most likely due to the noise in the satellite images, which makes it harder to discern influencing factors for air pollution.
The hand-labeled map images therefore help a lot as they already encode the desired entity labels as colors.

\paragraph{Experiment 4 --- OpenStreetMap and Google Maps Satellite}
The last experiment combines map and satellite imagery by concatenating both three-channel RGB images to one six-channel tensor.
OpenStreetMap images are used for the map images, since they have shown better performance than Google Maps in our task.
Satellite imagery is taken from Google Maps Satellite.
As both image sources use the same spatial resolution of \SI{80}{\meter} by \SI{80}{\meter}, local OpenStreetMap data should only be augmented by the satellite images.
However, the results in Table~\ref{tab:results} show no performance gain compared to using OpenStreetMap only.
In fact, the results are worse and significantly different for both models.

\begin{table}
    \centering
    \ra{1.3}
    \caption{\textbf{Results of baseline methods and experiments.} \model{} with OpenStreetMap images is providing the best performance overall, beating all baselines and all other \model{} variants. Using satellite images from Google Maps instead of OpenStreetMap images decreases the metric scores on the evaluation set. Combining both image types does not improve the score. The Random Forest model is outperforming all other baseline models on this data set which makes it the best baseline. All results are significantly different to each other.}
    \begin{tabular}{@{}lcc@{}}
    \toprule
    Model         & {\rsq} & {RMSE [\si{\micro\gram\per\cubic\meter}]}   \\ 
    \midrule
    \baselines
    \midrule
    \model{} experiment 1: OpenStreetMap     & \bfseries 0.673  & \bfseries 8.002  \\
    \model{} experiment 2: Google Maps & 0.537 & 8.918 \\
    \model{} experiment 3: Google Maps Satellite & 0.206 & 12.389 \\
    \model{} experiment 4: OpenStreetMap and Google Maps Satellite & 0.660 & 8.112 \\
    \bottomrule
    \end{tabular}
    \label{tab:results}
\end{table}

\paragraph{Computation times}
The computation times for the various models evaluated vary due to the different model complexities.
For example, with an Intel Xeon E5-2690V4 CPU training the MLP takes on average \num{25} seconds, Random Forest trains on average for \num{11} seconds, and the linear regression model is typically built in a single second.
Estimating pollution concentrations with these simple trained models for an area like Central London only takes seconds.
\model{} is the most complex model among them, but can be trained in reasonable time on commodity hardware.
We found that training it on a single consumer graphics card (Nvidia GTX 1080 TI) takes about \num{50} minutes.
Once \model{} is trained it can estimate pollution concentrations for the complete Central London area in \num{35} seconds.

\subsection{Model Analysis}

After seeing that \model{} can work well, we now further analyze the model.
First, we assess the data requirements of \model{}.
Then, we demonstrate that our model can be made interpretable by analyzing what our model has learned through guided backpropagation and by creating fake OpenStreetMap images.

\subsubsection{Analyzing Data Requirements}
\label{sec:transfer_learning}
The previous experiments showed that our model can successfully model air pollution in a data-driven way. 
While this is the main focus of this paper, the data we use (\num{3000} data points) is larger than those typically available in a real world setting.
In the following, we analyze the actual data requirements of all models evaluated above.
Overall, the corresponding results will inform future studies on data requirements and point towards necessary methodological advancements.

When gradually reducing the number of training data points, we noticed a drop in performance with smaller data sets for all models except for linear regression (cf. Figure~{\ref{fig:data_requirements}}).
Models based on neural networks experience a more pronounced performance loss in comparison to, for example, Random Forests, where there is only a slight decline.
We believe that this stems from the size and complexity of these models compared to simpler models like linear regression models.
More parameters need to be trained which tends to require more training examples.
However, about \num{30}\,\% of the training data is still sufficient for our model to exhibit performance comparable to the strongest baseline, which is the Random Forest trained on hand-crafted features.

Addressing the increased need for data is an important point for future work, which we also discuss in some more detail in Section~\ref{sec:discussion}.

\begin{figure}
    \centering
    \subfigure{\includegraphics[width=0.47\textwidth]{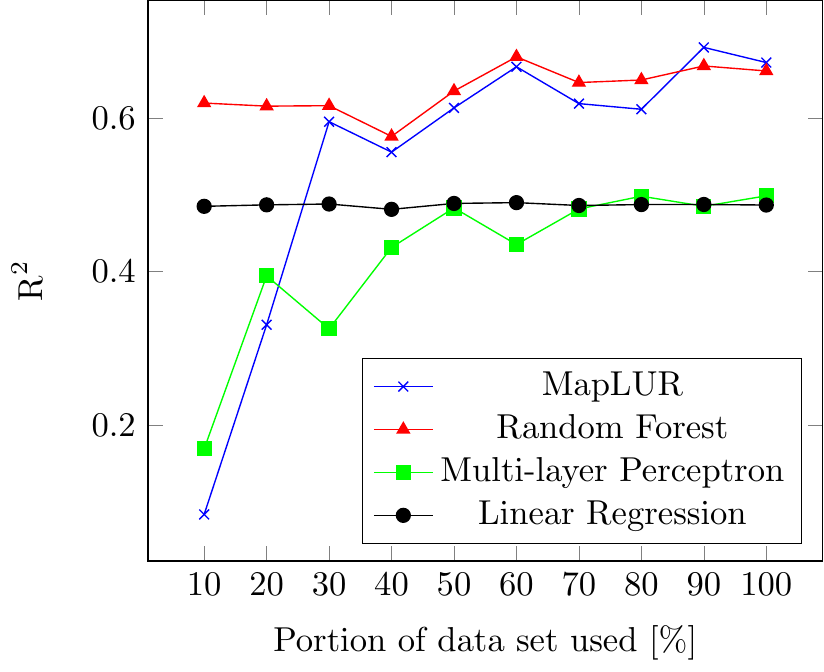}}
    \hfill
    \subfigure{\includegraphics[width=0.47\textwidth]{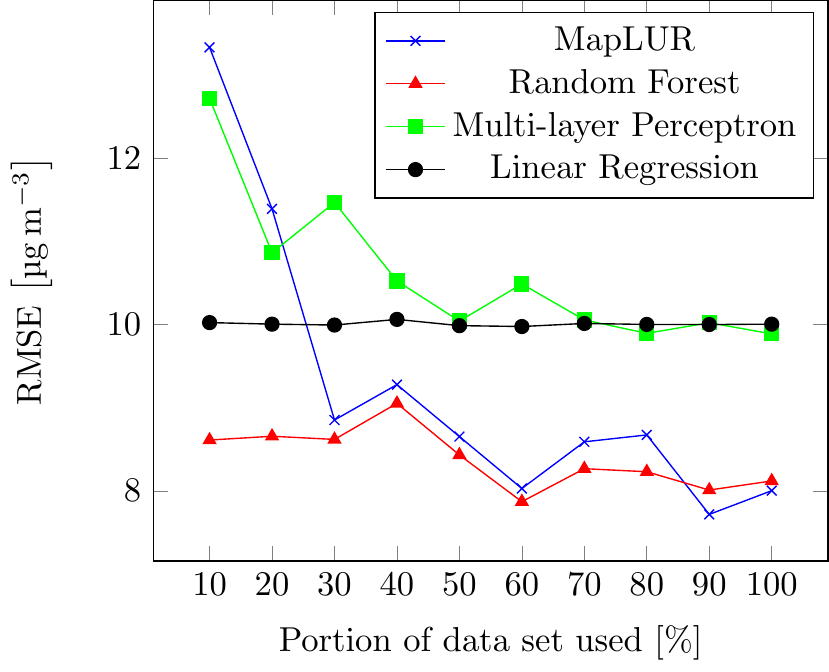}}
    \Description{
    The left graph shows \rsq{} values of \model{}, Linear Regression, Random Forest, and multi-layer perceptron models and the right graph shows RMSE values of these models for different training data set sizes. Both metrics are plotted in regards to the number of data points available for training. In both graphs, there is a sharp improvement in \model{}'s performance from {\num{20}}\,\% of the original data set to {\num{30}}\,\%. Its performances increases with more data. The multi-layer Perceptron starts with bad performance and also improves with more data. Linear Regression and Random Forests are relatively stable across all data set sizes. Random Forest shows slight gains with more data.}
    \caption{\textbf{Data requirements of land-use regression models.} 
    These graphs show the performance of commonly used land-use regression models and \model{} with varying training data set sizes. The portion of the data set used refers to the size of the \NOtwo{} data set with which the models were trained in our experiments. Thus, \num{100}\,\% is equivalent to \num{3000} data points. Each point in the graphs is the mean of \num{40} model runs, except for \model{}'s points which we only ran \num{5} times per data set size due to the model's computational complexity. This shows that \model{} can provide comparable results with \num{900} data points and it tends to improve with more data. Multi-layer Perceptrons behave similarly but they need more data to reach other baselines. Linear Regression and Random Forests are less dependent on data set size.}
    \label{fig:data_requirements}
\end{figure}

\subsubsection{Understanding Estimates using Guided Backpropagation}

In this section, we apply a technique called \emph{guided backpropagation}~\cite{springenberg2014striving}, which allows us to visualize the regions of the input image that the network focuses on for its estimation.
This approach starts off with a forward pass of an image.
Thereafter, the gradient of the activation is computed with respect to the input image.
At each ReLU in the model, positive gradients whose corresponding output during the forward pass was negative and negative gradients are set to \num{0} so that only features which contribute to the estimated pollution concentration are shown.
This allows us to visualize which parts of the image the model is paying attention to.
Several examples are shown in Figure~\ref{fig:guided_backprop}.

\begin{figure}
    \begin{tabular}{p{1.8cm}cccc}
      \vspace*{-1.5cm} Original images & \includegraphics[width=25mm]{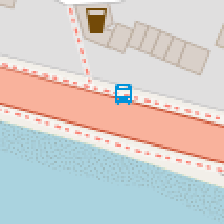} & \includegraphics[width=25mm]{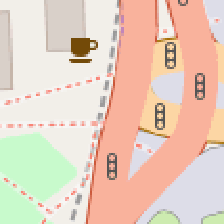} & \includegraphics[width=25mm]{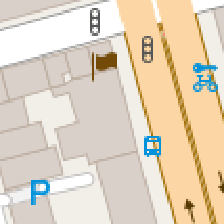} & \includegraphics[width=25mm]{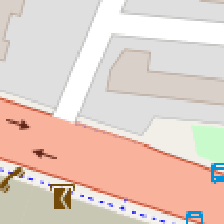} \\
      \vspace*{-1.5cm} Guided backpropagation &  \includegraphics[width=25mm]{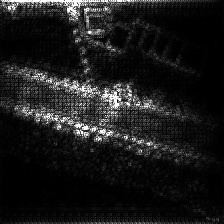} & \includegraphics[width=25mm]{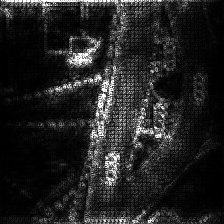} & \includegraphics[width=25mm]{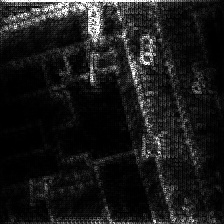} & \includegraphics[width=25mm]{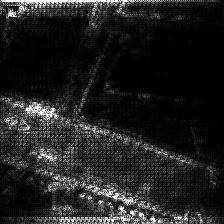} \\
    \end{tabular}
    \Description{There are four OpenStreetMap images with their respective guided backpropagation visualizations. The guided backpropagation images show that especially large streets contribute to the pollution estimate.}
    \caption{\textbf{Visualization of detected features using guided backpropagation~\cite{springenberg2014striving}.} This technique highlights important pixels in the input images by visualizing gradients of the activation with respect to pixel intensities. All negative gradients and positive gradients, whose corresponding output during the forward pass was negative, are set to \num{0} at each ReLU during backpropagation. This approach reveals parts of an image which contribute to the pollution. As one would expect, the model is concentrating on large streets. (Original images: \copyright~OpenStreetMap contributors)}
    \label{fig:guided_backprop}
\end{figure}

The guided backpropagation shows that the model is paying special attention to motorways, trunk roads, and primary roads which are rendered in red or orange colors in OpenStreetMap.
This shows that \model{} is able to automatically learn intuitively relevant features, since traffic is known as a large factor for \NOtwo{} pollution~\cite{carslaw2005estimations}.
\model{} also considers buildings, foot paths and cycle paths to some extent for its estimates while it seems to be ignoring water and park areas.
The model tends to pay more attention to pixels close to the center, which is understandable since we estimate the pollution concentration for the \SI{20}{\meter} by \SI{20}{\meter} areas that are in the center of each image.

\subsubsection{Analyzing Entity Influence using Artificial Map Tiles}

One of the biggest advantages of using a DOG-based model for air pollution estimation is that it extracts features by itself, while previous work always used hand-engineered features.
The leading question in developing land-use regression methods in previous work is: \emph{What entity} of \emph{what area} in \emph{what distance} to the center is contributing to the pollution?
From this, three categories of features arise: entity features, area features, and distance features.
We now want to investigate the correlation of these features with the model's output.
For this, we take advantage of the well-defined structure of map images with different color-coded entities.
Map images therefore can easily be recreated using graphic editing software, which makes it possible to create artificial OpenStreetMap images for which we can control the features separately while keeping all other features fixed.
We then observe changes to the model's output while modifying the values of these features.

\paragraph{Entity Features:} Entity features describe \emph{what} is seen on the image.
Entity features are often used for the estimation of air pollution, as, for example, industrial areas are usually contributing more to air pollution values than parks.
In this experiment, we investigate how certain entities are influencing the model estimate.
We build two kinds of images:
On the one hand, we create images that are each completely covered by one specific type of entity, resulting in uniformly colored square images.
On the other hand, the same images are then overlaid by the depiction of a motorway and a trunk road.
We expect that different underlying entities provide different estimates according to the usual presence of sources for \NOtwo{} pollution.
We also expect an increase in the air pollution estimate whenever a road is added to the underlying entity.
Depending on the type of road this increase might fluctuate.
Table~\ref{tab:feature_analysis_entity_features} shows the resulting pollution estimates by the CNN.

\begin{table}
    \centering
    \ra{1.3}
    \caption{\textbf{Model estimate for a given OpenStreetMap entity.} The entities span across the whole image and they are overlaid with different types of roads. Overlaying a road with another road does not make sense so these values are omitted. The `neutral' entity is a background that is used by OpenStreetMap for indicating land with no particular land-use. All estimates are in \si{\micro\gram\per\cubic\meter}.}
    \begin{tabular}{@{}c l l l l l l l l l@{}}
    \toprule
    Road type & \multicolumn{9}{c}{Entity Name} \\
\cmidrule{2-10} & \vtop{\hbox{\strut industrial}\hbox{\strut area}} & \vtop{\hbox{\strut residen-}\hbox{\strut tial area}} & \vtop{\hbox{\strut commer-}\hbox{\strut cial area}} & park & forest & water & neutral & \vtop{\hbox{\strut motor-}\hbox{\strut way}} & trunk \\
    \midrule
    no road & 37.71 & 38.29 & 38.72 & 38.87 & 39.27 & 41.66 & 42.06 	& 47.23		& 80.63 \\
    trunk	& 61.70	& 50.94	& 59.73 & 64.14 & 57.94	& 59.62	& 58.62 	& --- 		& --- \\
    motorway & 60.04 & 48.48 & 64.18 & 46.05 & 54.21 & 53.45 & 55.00 	& --- 		& --- \\
    \bottomrule
    \end{tabular}
    \label{tab:feature_analysis_entity_features}
\end{table}

Different underlying entities do not lead to large differences in pollution estimates if there is no road.
Only completely covering the image by a motorway or trunk road results in an estimate of over \SI{45}{\micro\gram\per\cubic\meter}.
Additionally, trunk roads seem to have a much higher impact on the air pollution estimate than motorways.
Adding a trunk road or motorway to any entity increases the air pollution estimate as expected.
The amount of increase depends on the underlying entity of the map and what kind of road is present.
This shows that the relationship of the entities that are visible in the map image are also important.
There seem to be complex correlations between different entity features, which cannot be modeled easily in simpler models like linear regression.

\paragraph{Area Features:} Area features describe \emph{how large} a given entity is in the image.
The area that an OpenStreetMap entity has on an image should contribute to the estimated pollution value.
In this experiment, we use the trunk road entity to show the influence of the area.
We build multiple images that contain a straight road that goes top to bottom or left to right through the center of the image.
As the background we always use the same neutral background that depicts general land-use in OpenStreetMap.
We then vary the width of that street either horizontally or vertically, depending on the street direction.
A linear increase in the street's width is equivalent to a linear increase in the street's area.

Figure~\ref{fig:feature_analysis_area_features} shows some of the artificial OpenStreetMap images as well as a plot of \model{}'s output given the street width in pixels.
As expected, an increasing width --- and therefore an increasing area --- of the street tends to increase the pollution estimate.
Both horizontal and vertical growth have very similar curves that are not linear but instead seem to be more logarithmic.
The similarity was expected, as during training, the images are augmented by rotation and flipping such that the direction of streets should not have any impact on the overall output.

\begin{figure}
    \subfigure[Two fake examples with a trunk road of width 10 px and 60 px, respectively.]{
    \begin{minipage}[b]{0.18\textwidth}
    \includegraphics[width=\textwidth]{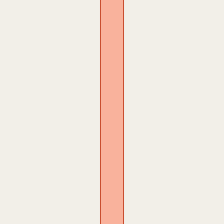}\\
    \includegraphics[width=\textwidth]{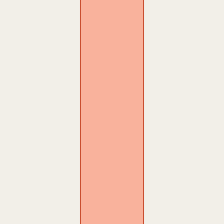}
    \end{minipage}
    }
    \hfill
    \subfigure[\NOtwo{} estimate for a given width of the road.]{
    \includegraphics[width=0.75\textwidth]{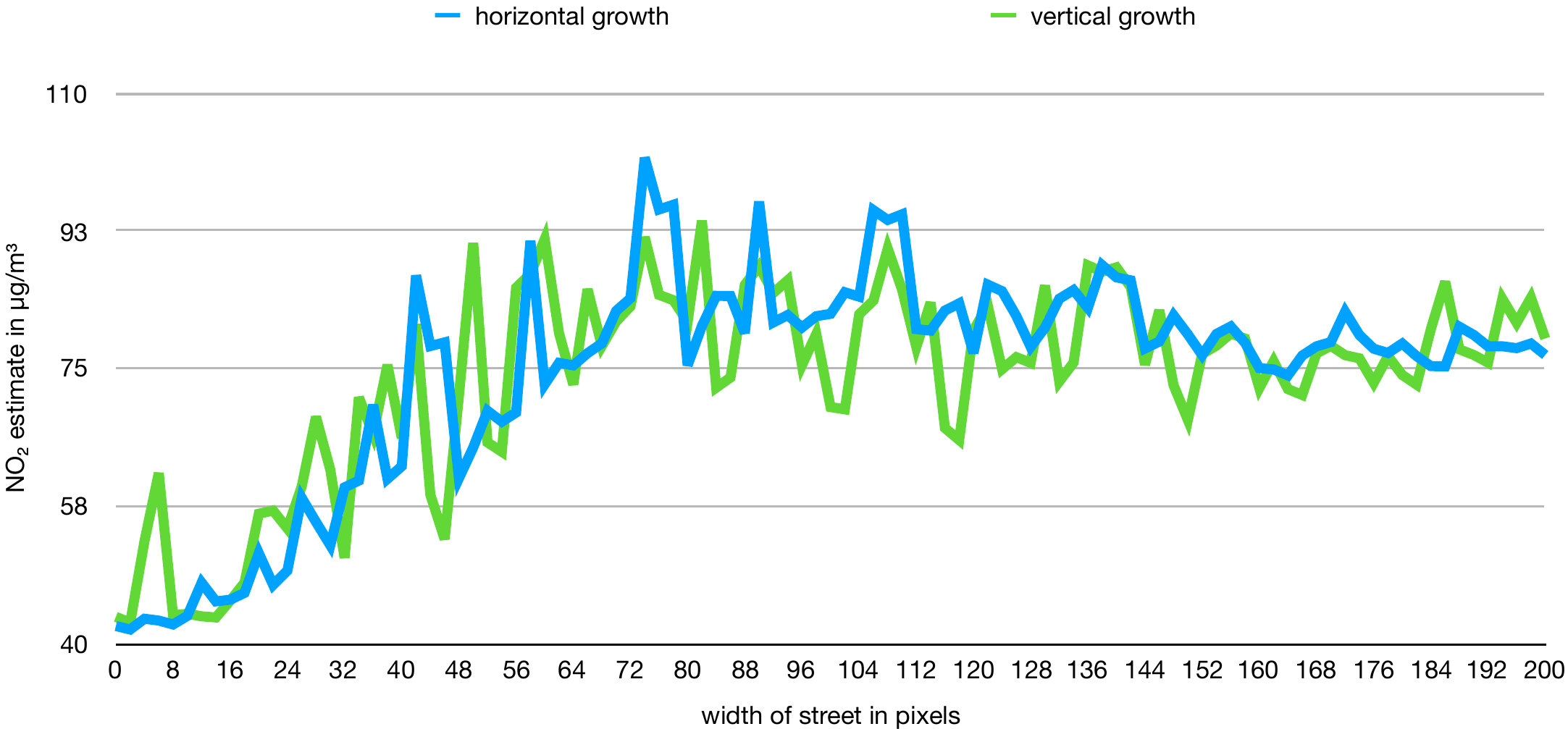}}
    \Description{Left are example images which show two fake OpenStreetMap images with a trunk road on each of them but with road widths. The line plot on the right shows how the pollution estimate is changing when the width of the trunk road is changed. When the road is only a few pixels thin, the estimate is slightly over \SI{40}{\micro\gram\per\cubic\meter} but when the road is more than 70 px wide, the estimate is typically over \SI{75}{\micro\gram\per\cubic\meter}.}
    \caption{\textbf{Varying the width/area of the street while keeping other features such as distance to the center and type of street fixed.}}
    \label{fig:feature_analysis_area_features}
\end{figure}

\paragraph{Distance Features:} Distance features describe \emph{how far away} a given entity is from the image center.
For this experiment, we create images that contain only one straight trunk road that is then moved vertically or horizontally, depending on the direction of the street.
With this setup, we can control the distance of the motorway to the center of the image while fixing the area and entity features.
We expect that the model produces higher estimates for images where the street is closer to the center, as this behavior was already observed in the guided backpropagation results.
Also, the desired value from LAEI is coming from a \SI{20}{\meter} subframe of the image which is in the image center.
The model therefore should have learned a tendency to weight features from the center of the image more than from the borders.

Figure~\ref{fig:feature_analysis_distance_features} shows image samples and the resulting estimation curves when moving the trunk road farther away from the center of the image.
As expected, the proximity of a street to the image's center contributes to the overall \NOtwo{} estimate positively.
Pearson correlations of the distance with the estimated values are always lower than -0.6, indicating a relatively strong negative correlation.
The curves that are shown are also not linear and can be better fitted by polynomials with a squared feature term than by a line.
To capture this non-linearity, more sophisticated methods need to be used, which justifies the use of Random Forests or neural networks.

\begin{figure}
    \subfigure[Two fake examples with a trunk road that is moved by \SI{10}{px} and \SI{60}{px}, respectively.]{
    \begin{minipage}[b]{0.18\textwidth}
    \includegraphics[width=\textwidth]{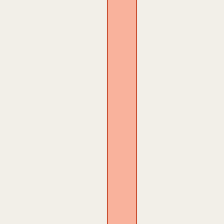}\\
    \includegraphics[width=\textwidth]{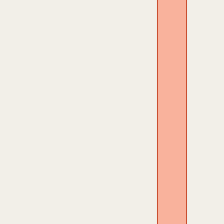}
    \end{minipage}
    }
    \hfill
    \subfigure[\NOtwo{} estimate for a given distance from the center of the image to the trunk road.]{
    \includegraphics[width=0.75\textwidth]{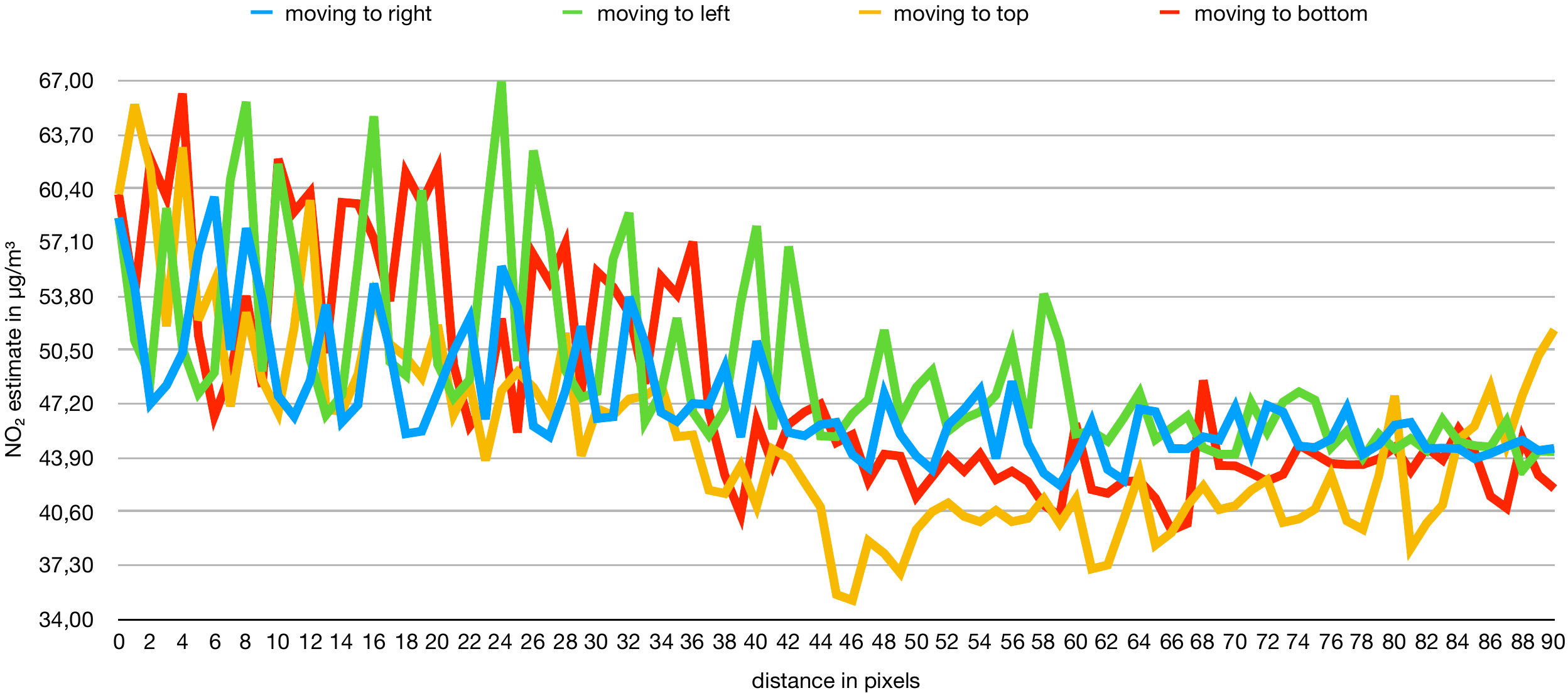}}
    \Description{Left are example images which show two fake OpenStreetMap images with a trunk road on each of them but with varying distances to the image's center. The line plot on the right shows how the pollution estimate is changing when the trunk road is moved away from the center of the image. When the road is in the middle, the estimate is around \SI{60}{\micro\gram\per\cubic\meter} but when the road is moved \SI{50}{\px} away the estimate is typically only around \SI{45}{\micro\gram\per\cubic\meter}.}
    \caption{\textbf{Varying the distance of a trunk road to the image center pixel by pixel while keeping other features fixed.}}
    \label{fig:feature_analysis_distance_features}
\end{figure}

\section{Discussion}
\label{sec:discussion}
In this section, we discuss some advantages and some current limitations of our proposed paradigm and model.
We believe that the advantages provide valuable additions to the models currently applied in land-use regression.
Since this is the first paper applying a model based on our new paradigm, there are still some limitations regarding the applicability of our model to real world data sets, for which we provide some possible ways to overcome.

\paragraph{Interpretability}
Firstly, purely data-driven models tend to be harder to interpret than simpler models, which is why they are often thought of as black-box models.
This also raises the concern that the models may put too much focus on unreliable features that explain the specific data set well, but fail to generalize to other data sets.
To alleviate these concerns, we have shown  that it is possible to reveal the inner workings of the model,
finding that the model's output heavily relies on land-use features such as streets or commercially used areas.
These features are also commonly employed by traditional land-use regression models.
We have also shown that \model{} implicitly focuses on other commonly used land-use information such as distance and area features.
This illustrates that purely data-driven approaches can yield interpretable models.

\paragraph{Feature extraction and complexity}
In addition to information that closely resemble hand-crafted features, our model is able to extract signals from the image in an automated and optimized fashion.
Thus, it can potentially capture more complex signals than modeled by hand-crafted features.
For example, it is likely that features like curvature of streets, street signs, and traffic lights are also considered by \model{} to estimate the air pollution.
However, an analysis of these features remains future work.
In particular, we believe that applying further model analysis techniques will enable researchers to find previously unknown features that can then be evaluated by experts and transferred to other, traditional land-use regression models.
Thus, \model's ability to extract interpretable features in combination with its inherent potential to model more complex relations of land-use and air pollution makes it a powerful tool for land-use regression.

\paragraph{Overfitting and generalization}
Despite our strong results on the LAEI data set and gaining an intuition for how \model{} works, 
we were not able to test generalizability across areas of interest due to the lack of similar data sets on different cities.
While our setup allows for generalizability in principle, in practice, certain challenges may arise.
In particular, deep learning methods are prone to overfitting, i.e, they may underperform when applied to input data that is very dissimilar to or not covered by the training data~\cite{demuth2014neural}.
However, we suspect that this problem is less pronounced for \model{} since map images are very structured and the model is therefore likely to see the vast majority of entity types that exist in the study area during training.
Additionally, we want to stress again that previous models are often not applicable to new areas at all, since they often rely on data that is either not open or only locally available.
Nonetheless, the model needs to be evaluated for every new application area before relying on its estimations. 
In order to use the model in new areas, it will usually be necessary to fit the model to some data from this area.
Therefore, it is important to have training, validation, and test data sets that are similar in characteristics and representative of the whole study area.
Typically, a simple random split is enough to achieve this~\cite{demuth2014neural}.

\paragraph{Data Requirements and Application to Real World Data}
\model{} uses a CNN that contains a large number of weights due to its architecture.
Training this deep-learning-based CNN requires more data than regular land-use regression models, which is why we have evaluated the approach on data from a model~\cite{laei} instead of real world data.
We have shown that \model{} works well given the \num{3000} data points of LAEI's modeled \NOtwo{} concentrations.
Since this number is far greater than most real world data sets for land-use regression, future work needs to
investigate the possibility of applying models based on the DOG paradigm in more realistic settings.
We believe that one promising approach for research in this direction is the use of transfer learning,
which has been shown to be an effective way of dealing with low-resource settings in both the areas of computer vision
\cite{mahajan2018exploring} and natural language processing \cite{howard2018universal}.
Transfer learning could be applied to MapLUR or other models based on the DOG paradigm by pre-training the model on a
large data set, like for example the LAEI data, and then fine-tuning it to a smaller data set of real world
measurements.
The global nature of features used in models based on the DOG paradigm ensures that this approach is generally possible.
Additional large-scale data sets can also be collected in the context of mobile measuring campaigns~\cite{hasenfratz2014pushing,montagne2015land}.
These data sets can then be used to provide further training data for the pre-training of DOG-based models.

\paragraph{Incorporating Distant Sources of Pollution}
Our analysis has shown that using images depicting \SI{80}{\meter} by \SI{80}{\meter} areas for each data point leads to good results, as can be seen in Appendix~\ref{sec:appendix_area_size}.
However, previous approaches to air pollution estimation based on land-use regression have shown that it is useful to
include information from wider surrounding areas in their features \cite{ryan2007review}.
In the specific case of the \model{} model, the size of the surrounding area that can be used is bounded by the
resolution of the input images:
If the area gets too large, the resolution of 224 x 224 px is not sufficient to encode the corresponding image.
While this could be countered by increasing the input resolution, this would significantly increase the cost of
training and prediction.
Therefore, it is an interesting direction for further research to develop models that can take into account larger
surrounding areas without increasing the image resolution.
This could for example be achieved by using stacked convolutional neural networks or a combination of convolutional and recurrent neural networks.

\paragraph{Integration of Additional Data Sources}
While we have focused only on map images as input, \model{} was able to outperform all considered baselines.
Nevertheless, previous work has shown that additional information can greatly improve the performance of air pollution models.
One example of such data would be elevation maps~\cite{ryan2007review}, which can be
integrated into \model{} in a way similar to Experiment~\num{4},
where we provided the CNN with additional map image layers.
Beyond this, there is a wide variety of methods to provide deep learning methods with additional information which holds great potential to further improve our results~\cite{mcauley2012image,aralikatte2019rewarding}.

\section{Conclusion}
\label{sec:conclusion}
In this paper, we have advocated DOG, a solely data-driven paradigm for air pollution estimation through land use regression.
Models that follow this paradigm do not require manually engineered features and are based on data that is openly and globally available.
This will ultimately result in models that are globally generalizable and can be applied in any area without modification.
Working towards this goal, we have presented \model{}, a deep learning based land-use regression model for air pollution estimation.
We have shown that it can estimate \NOtwo{} concentrations better than all considered baselines on a data set of modeled data from the Greater London area.
While our analysis of \model{} has shown that its data requirements are higher than commonly available data set sizes, we argued that transfer learning is a promising approach to alleviate this issue. 
We have also explored ways to analyze the factors that influence the prediction of this model, finding that a data-driven model architecture can be made interpretable by careful inspection of the trained model.
Thus, overall, this paper demonstrates the feasibility and advantages of our proposed data-driven paradigm DOG for land-use regression based air pollution modeling.

Future directions encompass work to further reduce the data requirements of data-driven models, the development of a comprehensive framework for extracting and interpreting features, as well as in-depth studies on real-world data, large-scale mobile measurements, and different cities.

\begin{acks}
This work has been partially funded by the DFG grant “p2Map: Learning Environmental Maps - Integrating Participatory Sensing and Human Perception”.
\end{acks}

\bibliographystyle{ACM-Reference-Format}
\bibliography{references}

\clearpage
\appendix

\section{Sampled LAEI Cells}
\label{sec:appendix_laei_cells}
Figure~\ref{fig:data_split} depicts the cells which we sampled from LAEI.
The blue cells are used for training our models while the red cells are used to evaluate model performance for unseen locations.

\begin{figure}
    \includegraphics[width=0.9\textwidth]{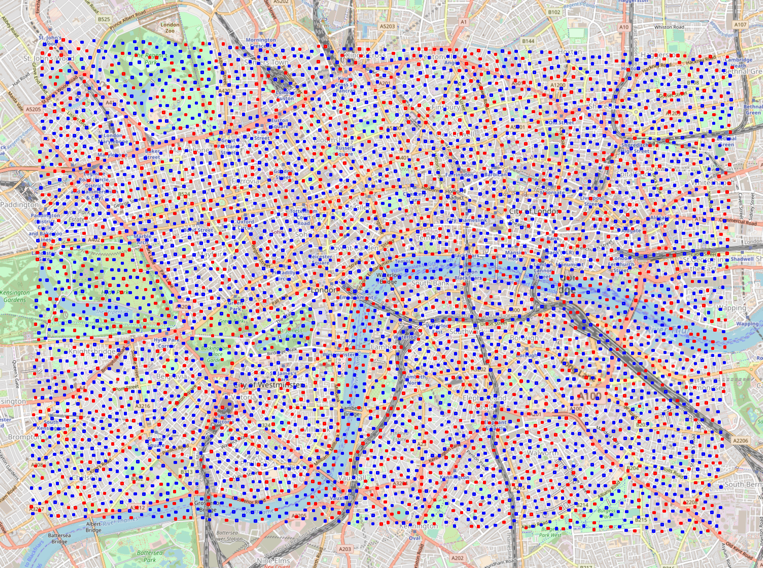}
    \Description{A map where the locations of the LAEI cells, which are used for model training and evaluation, are shown.}
    \caption{\textbf{LAEI cells sampled for the experiments.} The \num{3000} blue cells are the training data set and the \num{1500} red cells are the test data set. (Underlying OpenStreetMap Image: \copyright~OpenStreetMap contributors)}
    \label{fig:data_split}
    \end{figure}

\section{Analysis Of Area Size}
\label{sec:appendix_area_size}
In this work, we only used \SI{80}{\meter} by \SI{80}{\meter} images as inputs for our model.
However, despite the potential evaluation issues with overlapping images described in Section~\ref{sec:air_pollution_data}, it is still interesting to see how our model behaves when it is able to see more or less of the surroundings.
Therefore, the model was provided with OpenStreetMap images depicting square areas around the data point with side lengths of \SI{60}{\meter}, \SI{80}{\meter}, \SI{100}{\meter}, \SI{200}{\meter}, \SI{500}{\meter}, and \SI{1000}{\meter} while maintaining a resolution of 224 px by 224 px.
The mean results after \num{40} evaluations can be seen in Table~\ref{tab:larger_areas}.

\begin{table}
    \centering
    \ra{1.3}
    \caption{\textbf{Evaluating the influence of the area size depicted in each map image on \model{}'s performance.} The results suggest that \SI{80}{\meter} by \SI{80}{\meter} areas are optimal.}
    \begin{tabular}{@{}lcc@{}}
    \toprule
    Model         & {\rsq}  & {RMSE}   \\
    \midrule
    \baselines
    \midrule
    \SI{60}{\meter}      & 0.626  & 8.511  \\
    \SI{80}{\meter}      & \bfseries 0.673  & \bfseries 8.002  \\
    \SI{100}{\meter}      & 0.637  & 8.381  \\
    \SI{200}{\meter}      & 0.618  & 8.603  \\
    \SI{500}{\meter}      & 0.597  & 8.833  \\
    \SI{1000}{\meter}      & 0.390  & 10.856  \\
    \bottomrule
    \end{tabular}
    \label{tab:larger_areas}
\end{table}

Our model does not benefit from the increased image size as can be seen from both \rsq{} and RMSE.
The mean performance decreases consistently with each increase in depicted area size over \SI{80}{\meter} by \SI{80}{\meter}.
This implies that the potential evaluation issue with overlapping images, which is described in Section~\ref{sec:air_pollution_data}, is not very severe since the model should be gaining performance with larger images otherwise.
It also suggests that the very close surroundings are important and that information from further away is not helping.
However, the model is suffering from performance loss with smaller areas than \SI{80}{\meter} by \SI{80}{\meter}.
Thus, it seems that a side length of \SI{80}{\meter} for the depicted square areas is optimal for \model{} especially considering the fact that all other results are significantly different according to the Wilcoxon signed-rank test even after Bonferroni correction, since $p < \frac{0.05}{n}$ for $n = 9$.

\section{Evaluation Metrics}
\label{sec:appendix_evaluation_metrics}
Given the desired target values $y = \{y_1, y_2, \dots, y_n\}$ and the model's
output $\hat{y} = \{\hat{y}_1, \hat{y}_2, \dots, \hat{y}_n\}$, two commonly used
metrics in land-use regression papers, namely \rsq{} and root-mean-square error
(RMSE)~\cite{champendal2014air,liu2015land,eeftens2012spatial}, are used to
evaluate the model on the evaluation set.

On the one hand, \rsq{} describes how much of the target's variation is explained by the model:

\[ \text{R}^2(y, \hat{y}) = 1 - \frac{\sum_{i=1}^{n} (y_i - \hat{y}_i)^2}{\sum_{i=1}^{n} (y_i - \bar{y})^2}, \]

where $\bar{y} = \frac{1}{n} \sum_{i=1}^{n} y_i$ is the mean of all desired target values.
The metric can take values from $-\infty$ to \num{1}.
A \rsq{} of \num{1} indicates a perfect fit.
A value of \num{0} is achieved by always estimating the mean of the evaluation set's target values.
Negative values indicate that the model is worse than always estimating the mean.

On the other hand, the RMSE is, as the name already suggests, the square root of the mean of the squared errors:

\[ \text{RMSE}(y, \hat{y}) = \sqrt{\frac{1}{n} \sum_{i=1}^{n} (y_i - \hat{y_i})^2}. \]

Thus, RMSE can only take non-negative values, where \num{0} would be perfect for this metric and larger RMSEs are worse.

\section{Model Building Procedures For The Baselines}
\label{sec:appendix_model_building}
The following explains the model building procedures for the baseline methods in more detail.

\subsection{Linear Regression}
\label{sec:appendix_model_building_linear_regression}
The first baseline model we consider is the commonly used linear regression.
The model's development is based on a supervised stepwise selection which was
used in the Escape project~\cite{eeftens2012development} for land-use regression
model development before.
Each predictor variable is ranked based on the model's adjusted \rsq{} from a univariate regression.
The adjusted \rsq{} used by~\citet{eeftens2012development} is like the \rsq{} but penalizes adding variables which do not fit the model.
Thus, ideally only independent variables which affect the dependent variable are used.
If there are variables that are of the same category but with different buffer sizes, then only the variable with the highest score is considered for use in the final model due to the high correlation of these variables between each other.
The model starts with the variable that achieved the highest score.
Thereafter, each one of the remaining variables is temporarily added to the model, evaluated, and the best performing variable is added to the model permanently if it increases the model's adjusted \rsq{} by at least 0.01.
This is repeated until no variables are left.
Then all selected variables with a p-value greater than 0.1 are removed and the resulting model is fit again, just like described in~\citet{eeftens2012development}.
Finally, the variance inflation factors (VIFs) are calculated for each variable in order to quantify the increase in variance due to collinearity of the variables.
If a variable has an VIF that is greater than \num{3}, the variable with the largest VIF is removed and the model is refit.
In accordance with~\citet{eeftens2012development} this is also repeated until no variable has an VIF greater than \num{3}.

\subsection{Random Forest}
\label{sec:appendix_model_building_random_forest}

\begin{table}
    \centering
    \ra{1.3}
    \caption{\textbf{Random Forest hyperparameters optimized using stochastic search with the corresponding search spaces.}}
    \begin{tabular}{@{}lc@{}}
    \toprule
    Hyperparameter & Search space \\
    \midrule
    Number of trees & 1 to 1000\\
    Fraction of features to consider at most per split & 0.0 to 1.0 \\
    Minimum samples required to be a leaf node & 1 to 100 \\
    Minimum samples required to split a node & 2 to 20\\
    Build trees with bootstrap samples & True or False \\
    \bottomrule
    \end{tabular}
    \label{tab:rf_hyperparams}
\end{table}

Another baseline is a Random Forest model which employs ensembles of decision trees for its estimations~\cite{breiman2001random}.
This model is built in a similar way to the procedure
in~\citet{brokamp2017exposure}.
The steps of the model building procedure are described in the following.
First, the best buffer radii for each type of variable were determined based
on the adjusted \rsq{} of a univariate regression on the training data set.
As described in Section~\ref{sec:appendix_model_building_linear_regression} before, the adjusted \rsq{} is like the \rsq{} but penalizes adding variables which do not fit the model.
Then, an initial Random Forest is considered for the following which uses
values that have shown to work decently in preliminary experiments:
It builds \num{500} trees, considers half the available features when looking
for the best split, and uses the default values of scikit-learn's Random
Forest implementation for the other hyperparameters~\cite{scikit-learn}.
The best variables of each type are then fitted using this Random Forest to rank
the variables based on the variable importance score of the Random Forest.
Thereafter, the least important variables are removed iteratively.
For each iteration, the model is fitted to the remaining variables on the
training data set and the out of bag \rsq{} is calculated by estimating each training sample without using the trees that had the training sample in their bootstrap sample.
This metric can be used with Random Forests to estimate performance without an independent test set.
The set of variables that achieved the best performance is selected for further
use.
Then the best hyperparameters for the Random Forest are found by a stochastic
search that samples hyperparameter values and evaluates them with a ten fold
cross-validation on the folds of the training set.
Each hyperparameter value is sampled from a uniform distribution with a specific search space.
Table~\ref{tab:rf_hyperparams} shows these hyperparameters with their search spaces.
This search ran for three hours and used six CPU cores of an Intel Xeon E5-2690V4 processor to fit as many models as
possible.
During this search, \num{1051} sets of hyperparameters were evaluated.
The best hyperparameters are then used to fit a Random Forest on the complete
training set and evaluate the model on the test set.

\subsection{Multi-layer Perceptron}
\label{sec:appendix_model_building_mlp}
For the last baseline, we employ a multi-layer Perceptron, which is a type of neural network.
For this model, we base our model building procedure on the one used by~\citet{alam2015exploring}.
We adapt it slightly by first selecting the best buffer radius for each variable type based on the adjusted \rsq{} of an univariate regression.
We use the adjusted \rsq{} again for this selection to be consistent with the model building procedures for linear regression and Random Forest described before.
Then we search for the best performing architecture by evaluating models with different number of hidden layers and neurons for each layer with ten fold cross-validations on the training data set.
This is done by randomly sampling the number of layers from a uniform distribution ranging from \num{1} to \num{3} hidden layers and randomly sampling the number of neurons for each layer from another uniform distribution which ranges from \num{1} to \num{30}.
These bounds are chosen since they encompass the architectures of all previously published models that we found~\cite{buevich2016modeling,adams2015advancing,alam2015exploring,liu2015land,champendal2014air}.
We randomly sample model architectures and evaluate them for \num{24} hours during which \num{2629} architectures were evaluated.
The best performing architecture is then trained on the complete training set and evaluated on the test set.

\section{Comparison Of OpenStreetMap And Google Maps Images}
\label{sec:appendix_comparison_osm_gmaps}
Figure~\ref{fig:comparison_osm_gmap} compares Google Maps and OpenStreetMap images, showing the differences between both default styles shown on the services' websites.
While OpenStreetMap provides at least four colors to denote different types of streets, Google Maps only uses two.
Google Maps does not color-code all of the available information to ease the visual effort of the user.
This however is not helpful for the CNN model, which works better with clear visual cues that denote entities.

\begin{figure}
    \centering
    \subfigure[OpenStreetMap images (\copyright~OpenStreetMap contributors)]{\includegraphics[width=0.39\textwidth]{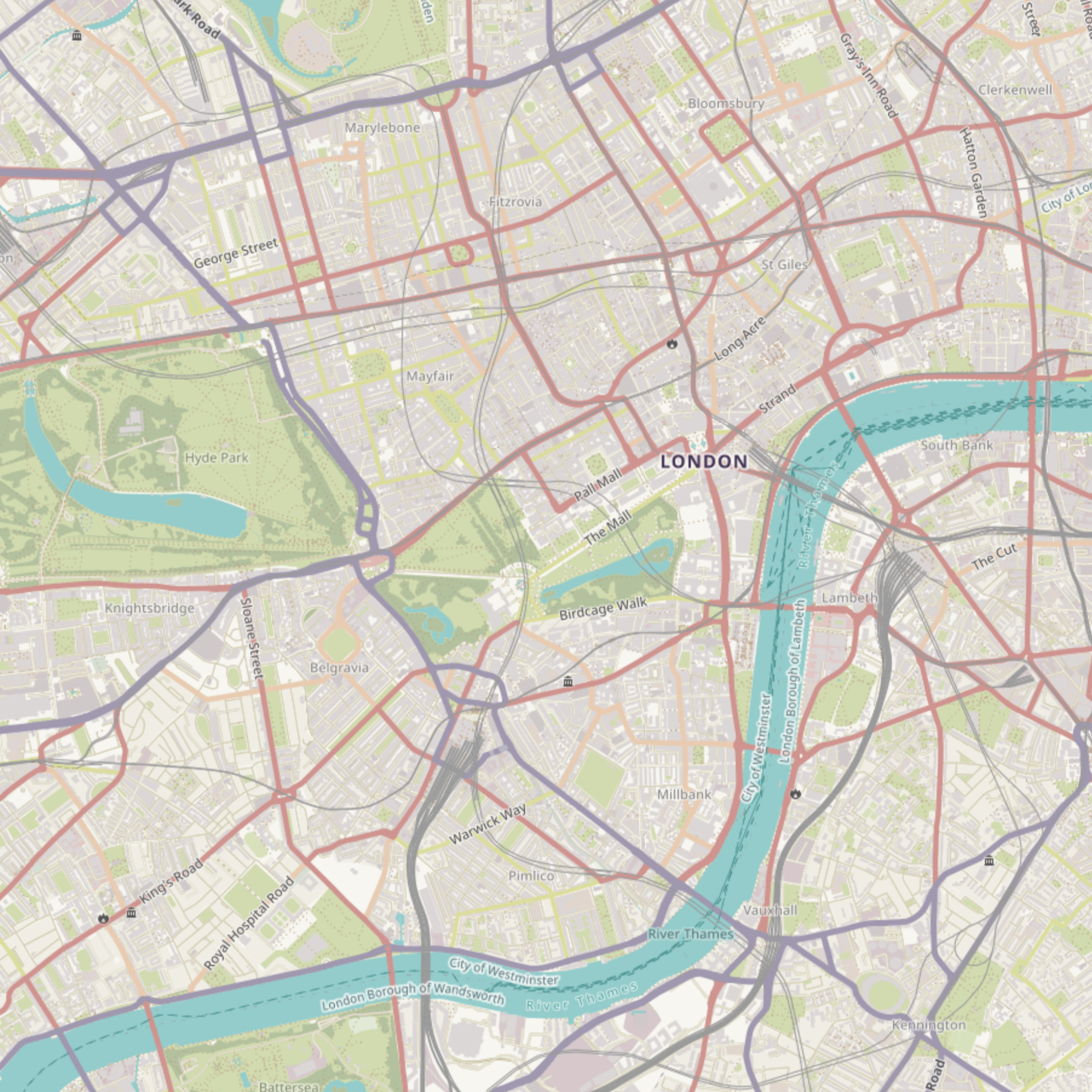}}\hspace{12mm}
    \subfigure[Google Maps images (Map data: Google)]{\includegraphics[width=0.39\textwidth]{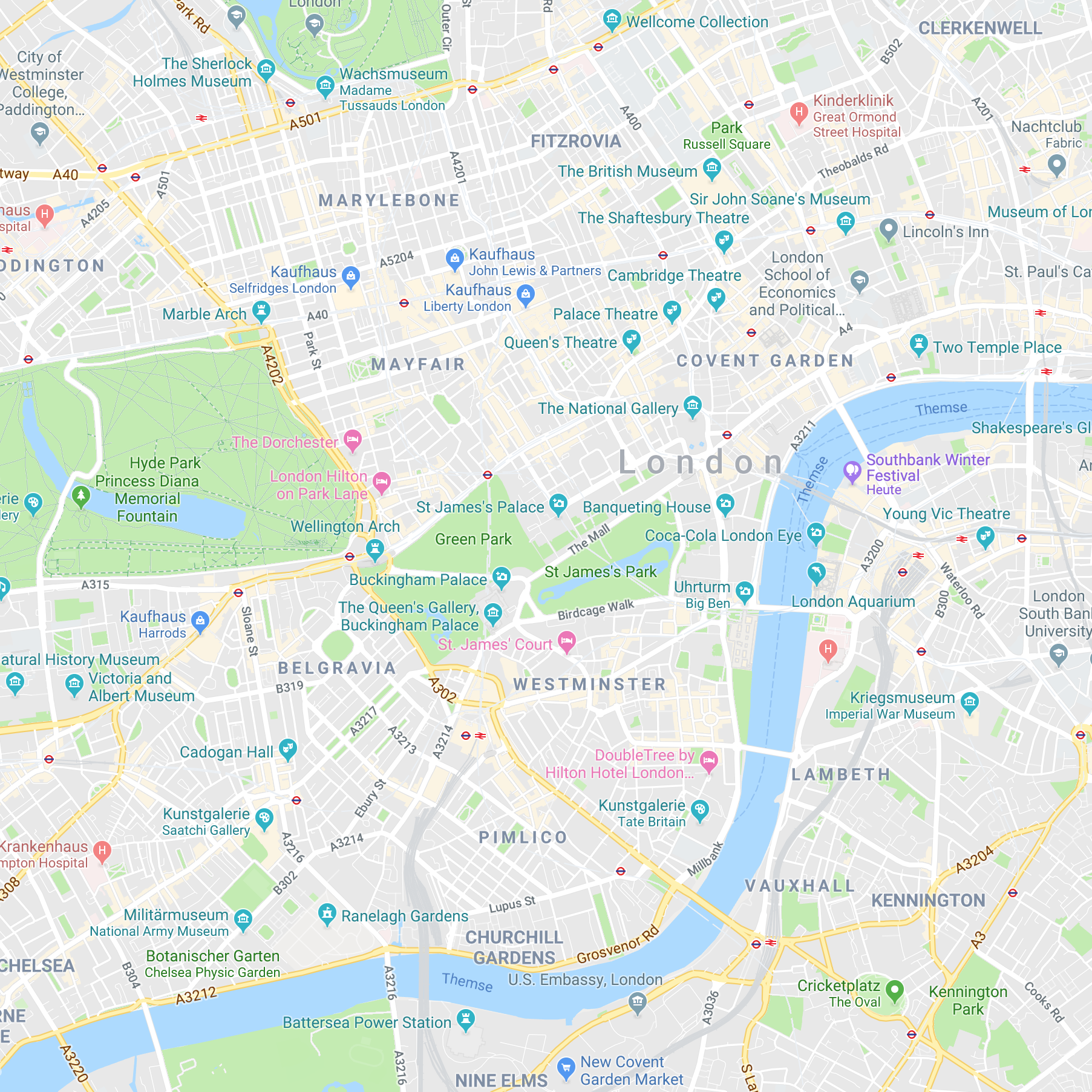}}
    \Description{Example of an OpenStreetMap image and an Google Maps image.}
    \caption{\textbf{Comparison of OpenStreetMap and Google Maps images.} OpenStreetMap images contain more color-coded entities such as streets. As the visual style is highly important for the CNN to learn the task, the simpler Google Maps style produces worse results.}
    \label{fig:comparison_osm_gmap}
\end{figure}

\end{document}